\documentclass[11pt,a4paper]{article}
\usepackage[hyperref]{acl2019}
\usepackage{times}
\usepackage{latexsym}
\usepackage{amssymb}
\usepackage{amsmath}
\usepackage{multirow}
\usepackage[pdftex]{graphicx}

\usepackage{url}

\aclfinalcopy 
\def\aclpaperid{751} 

 \DeclareMathOperator*{\argmax}{\arg\!\max}
 \newcommand{\hist}{\mbox{$\mathbf{y}_{1:i-1}$}}
\DeclareMathOperator{\E}{\mathbb{E}}

\title{Domain Adaptive Inference for Neural Machine Translation}
  
\author{Danielle Saunders$^\dag$ \and Felix Stahlberg$^\dag$ \and Adri\`a de Gispert$^{\ddagger}$ \and Bill Byrne$^{\ddagger\dag}$ \\
    $^\dag$Department of Engineering, University of Cambridge, UK  \\
    $^\ddagger$SDL Research, Cambridge, UK\\
      {\tt \{ds636, fs439, wjb31\}@cam.ac.uk}, {\tt \{agispert, bbyrne\}@sdl.com} \\
}

\begin{document}
\allowdisplaybreaks
\maketitle
\begin{abstract}
We investigate adaptive ensemble weighting for Neural Machine Translation, addressing the case of improving performance on a new and potentially unknown domain without sacrificing performance on the original domain. We adapt sequentially across two Spanish-English and three English-German tasks, comparing unregularized fine-tuning, L2 and Elastic Weight Consolidation. We then report a novel scheme for adaptive NMT ensemble decoding by extending Bayesian Interpolation with source information, and show strong improvements across test domains without access to the domain label.
\end{abstract}

\section{Introduction}
Neural Machine Translation (NMT) models are effective when trained on broad domains with large datasets, such as news translation \cite{bojar2017findings}.
However, test data may be drawn from a different domain, on which general models can perform poorly \cite{koehn2017six}.  We address the problem of adapting to one or more domains while maintaining good performance across all domains. Crucially, we assume the realistic scenario where the domain is unknown at inference time.

One solution is ensembling models trained on different domains \cite{freitag2016fast}. This approach has two main drawbacks. Firstly, obtaining models for each domain is challenging. Training from scratch on each new domain is impractical, while continuing training on a new domain can cause catastrophic forgetting of previous tasks \cite{french1999catastrophic}, even in an ensemble  \cite{freitag2016fast}. Secondly, ensemble weighting requires knowledge of the test domain. 

We address the model training problem with regularized fine-tuning, using an L2 regularizer \cite{barone2017regularization} and Elastic Weight Consolidation (EWC) \cite{kirkpatrick2017overcoming}. We fine-tune sequentially to translate up to three domains with the same model.

We then develop an adaptive inference scheme for NMT ensembles by extending Bayesian Interpolation (BI) \cite{allauzen2011bayesian} to sequence-to-sequence models.\footnote{See \texttt{bayesian} combination schemes at \url{https://github.com/ucam-smt/sgnmt}} This lets us calculate ensemble weights adaptively over time without needing the domain label, giving strong improvements over uniform ensembling for baseline and fine-tuned models.

\subsection{Adaptive training}
In NMT fine-tuning, a model is first trained on a task $A$, typically translating a large general-domain corpus \cite{luong2015stanford}. The optimized parameters $\theta^*_{A}$ are fine-tuned on task $B$, a new domain. Without regularization, catastrophic forgetting can occur: performance on task $A$ degrades as parameters adjust to the new objective. A regularized  objective is:
\begin{equation}
L(\theta) = L_B(\theta) + \Lambda \sum_j F_j (\theta_j - \theta^*_{A,j})^2
\label{eq:regularization}
\end{equation}
where $L_A(\theta)$ and $L_B(\theta)$ are the likelihood of tasks $A$ and $B.$
We compare three cases: 
\begin{itemize}
\item \textbf{No-reg}, where $\Lambda=0$
\item \textbf{L2}, where $F_j=1$ for each parameter index $j$
\item \textbf{EWC}, where $F_j=\E \big[ \nabla^2  L_A(\theta_j)\big] $, a sample estimate of task $A$ Fisher information. This effectively measures the importance of $\theta_j$ to task $A$. 
\end{itemize}

For L2 and EWC we tune $\Lambda$ on the validation sets for new and old tasks to balance forgetting against new-domain performance.

\subsection{Adaptive decoding} 

We extend the BI formalism to condition on a source sequence, letting us apply it to adaptive NMT ensemble weighting. We consider models $p_k(\mathbf{y} | \mathbf{x})$ trained on $K$ distinct domains, used for tasks $t=1,\ldots,T$. In our case a task is decoding from one domain, so $T=K$.  We assume throughout that $p(t)=\frac{1}{T}$, i.e.\ that tasks are equally likely absent any other information.

A standard, fixed-weight ensemble would translate with:
\begin{equation}\argmax_\mathbf{y} p(\mathbf{y} |\mathbf{x}) = \argmax_\mathbf{y}  \sum_{k=1}^K W_k  p_k(\mathbf{y} | \mathbf{x})\label{eq:weight-ensemble}\end{equation}
The BI formalism assumes that we have tuned sets of ensemble weights $\lambda_{k,t}$ for each task.  
This defines a task-conditional ensemble
\begin{equation}p( \mathbf{y} | \mathbf{x}, t) =  \sum_{k=1}^K \lambda_{k,t} \; p_k(\mathbf{y} | \mathbf{x})\end{equation}
which can be used as a fixed weight ensemble if the task is known.
However if the task $t$ is not known, we wish to translate with:
\begin{equation}
\argmax_\mathbf{y} p(\mathbf{y} |\mathbf{x}) = \argmax_\mathbf{y}  \sum_{t=1}^T p(t, \mathbf{y} | \mathbf{x})
\label{eq:task-unknown-decode}
\end{equation}
At step $i$, where $h_i$ is history $\hist$:
\begin{align}
         p(y_i|h_i,\mathbf{x}) &= \sum_{t=1}^T p(t, y_i|h_i, \mathbf{x}) \notag \\ 
         & = \sum_{t=1}^T p(t|h_i, {\bf x}) \, p(y_i|h_i, t, {\bf x}) \notag  \\
          &= \sum_{k=1}^K  p_k(y_i|h_i, \mathbf{x})  \sum_{t=1}^T p(t|h_i, \mathbf{x})  \lambda_{k,t}  \notag \\
        &= \sum_{k=1}^K  W_{k,i} \, p_k(y_i|h_i, \mathbf{x})  
\label{eq:adaptdecode}
 \end{align}
This has the form of an adaptively weighted ensemble where, by comparison with Eq. \ref{eq:weight-ensemble}:
\begin{equation}W_{k,i} = \sum_{t=1}^T p(t|h_i, \mathbf{x}) \lambda_{k,t} \label{eq:weight-adaptive-ensemble}\end{equation}
In decoding, at each step $i$  adaptation relies on a recomputed estimate of the {\em task posterior}: 
\begin{equation}p (t|h_i, \mathbf{x}) = \frac{p(h_i|t,\mathbf{x}) p(t|\mathbf{x})} {\sum_{t'=1}^T p(h_i|t',\mathbf{x}) p(t'|\mathbf{x})}\label{eq:task-posterior}\end{equation}

\subsubsection{Static decoder configurations}

In static decoding (Eq.~\ref{eq:weight-ensemble}), the weights $W_{k}$ are constant for each source sentence $\bf x$.   BI simplifies to a  uniform ensemble when  $\lambda_{k,t} = p(t|\mathbf{x}) = \frac{1}{T}$.  This leads to $W_{k,i} = \frac{1}{K}$ (see 
Eq.~\ref{eq:weight-adaptive-ensemble}) as a fixed equal-weight interpolation of the component models.

Static decoding can also be performed with task posteriors conditioned only on the source sentence, which reflects the assumption  that the history can be disregarded and that $p (t|h_i, \mathbf{x}) = p( t | \mathbf{x})$.       In the most straightforward case, we assume that only domain $k$ is useful for task $t$: $\lambda_{k,t}=\delta_k(t)$   (1 for $k=t$, 0 otherwise). Model weighting simplifies to a fixed ensemble: 
\begin{equation}
W_k = p(k|{\bf x})
\label{eq:wIS}
\end{equation}
and decoding proceeds according to Eq.~\ref{eq:weight-ensemble}. We refer to this as decoding with an {\em informative source} (IS).

We  propose using $G_t$, an collection of n-gram language models trained on source language sentences from tasks $t$, to estimate $p(t|\mathbf{x})$:
\begin{align} 
 \label{eq:lmsentence}
p(t|\mathbf{x}) = \frac{p(\mathbf{x}|t)p(t)}{\sum_{t'=1}^Tp(\mathbf{x}|t')p(t')} =\frac{G_t(\mathbf{x})}{\sum_{t'=1}^T G_{t'}(\mathbf{x})} 
\end{align}
In this way we use source language n-gram language models to estimate $p(t=k|{\bf x})$ in Eq.~\ref{eq:wIS} for static decoding with an informative source.

\subsubsection{Adaptive decoder configurations}
For adaptive decoding with Bayesian Interpolation, as in Eq.~\ref{eq:adaptdecode}, the model weights vary during decoding according to Eq.~\ref{eq:weight-adaptive-ensemble} and Eq.~\ref{eq:task-posterior}.   We assume here that $p(t|\mathbf{x}) = p(t) = \frac{1}{T}$. This corresponds to the approach in \citet{allauzen2011bayesian}, which considers only language model combination for speech recognition. We refer to this in experiments  simply as BI.
A refinement is to incorporate Eq. \ref{eq:lmsentence} into Eq.~\ref{eq:task-posterior}, which would be Bayesian Interpolation with an informative source (BI+IS).  

We now address the choice of $\lambda_{k,t}$.   A simple but  restrictive approach is to take $\lambda_{k,t}=\delta_k(t)$.  We refer to this as {\em identity-BI}, and it embodies the assumption that only one domain is useful for each task.

Alternatively, if we have validation data  $V_t$ for each task $t$,   parameter search can be done to optimize $\lambda_{k,t}$ for BLEU over $V_t$ for each task.    This is straightforward but relatively costly.

We propose a  simpler approach based on the source language n-gram language models from Eq. \ref{eq:lmsentence}.  We assume that each $G_t$ is also a language model for its corresponding domain $k$. With $\overline{G}_{k,t} = \sum_{\mathbf{x} \in V_t} G_k(\mathbf{x})$, we take: \begin{equation}
    \lambda_{k,t} = \frac{\overline{G}_{k,t}}{\sum_{k'} \overline{G}_{k',t}}
    \label{eq:lambda-bisd}
\end{equation} 
$\lambda_{k,t}$ can be interpreted as the probability that task $t$ contains sentences $\mathbf{x}$ drawn from domain $k$ as estimated over the $V_t$.  
 
\begin{figure}[t!]
\centering
\includegraphics[width=\linewidth]{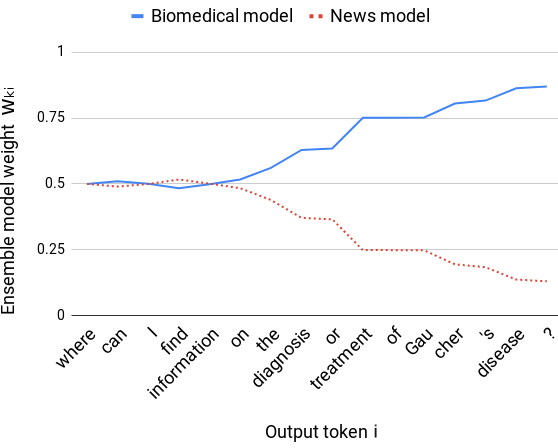}
\caption{Adaptively adjusting ensemble model weights $W_{k,i}$ (Eq. \ref{eq:weight-adaptive-ensemble}) during decoding with BI}
\label{fig:bi}
\end{figure}
Figure \ref{fig:bi} demonstrates this adaptive decoding scheme when weighting a biomedical and a general (news) domain model to produce a biomedical sentence under BI. The model weights $W_{k,i}$ are even until biomedical-specific vocabulary is produced, at which point the in-domain model dominates.

\subsubsection{Summary}
We summarize our approaches to decoding in Table \ref{tab:bi}.

\begin{table}[!ht]
\centering
\begin{tabular}{|l|l|cc | }
\hline
 &\textbf{Decoder}     & $\mathbf{p(t|x)}$   &  $\mathbf{\lambda_{k,t}}$\\
    \hline
\multirow{2}{*}{Static}&Uniform & $\frac{1}{T}$ & $\frac{1}{T}$ \\
&IS &Eq. \ref{eq:lmsentence}& $\delta_k(t)$ \\ \hline
\multirow{3}{*}{Adaptive}&Identity-BI &$\frac{1}{T}$  & $\delta_k(t)$ \\
&BI & $\frac{1}{T}$  &Eq. \ref{eq:lambda-bisd}\\
&BI+IS & Eq. \ref{eq:lmsentence}  &Eq. \ref{eq:lambda-bisd} \\
\hline
    \end{tabular}
\caption{Setting task posterior $p(t|\mathbf{x})$ and domain-task weight $\lambda_{k,t}$ for $T$ tasks under decoding schemes in this work. Note that IS can be combined with either Identity-BI or BI by simply adjusting $p(t|h_i, \mathbf{x})$ according to Eq. \ref{eq:task-posterior}.}

\label{tab:bi}
\end{table}

\subsection{Related Work}

Approaches to NMT domain adaptation include training data selection or generation \cite{sennrich2016improving, wang2017sentence, sajjad2017neural} and fine-tuning output distributions \cite{dakwale2017fine, khayrallah2018regularized}. 

\citet{vilar2018learning} regularizes parameters with an importance network, while \citet{thompson2018freezing} freeze subsets of the model parameters before fine-tuning. Both observe forgetting with the adapted model on the general domain data in the realistic scenario where the test data domain is unknown. \citet{barone2017regularization} fine-tune with L2 regularization to reduce forgetting. Concurrently with our work, \citet{thompson2019ewc} apply EWC to reduce forgetting during NMT domain adaptation. 

During inference, \citet{garmash2016ensemble} use a gating network to learn weights for a multi-source NMT ensemble. \citet{freitag2016fast} use uniform ensembles of general and no-reg fine-tuned models. 

\section{Experiments}
\label{ss:data}
We report on Spanish-English (es-en) and English-German (en-de). For es-en we use the Scielo corpus \cite{neves2016scielo}, with Health as the general domain, adapting to Biological Sciences (`Bio'). We evaluate on the domain-labeled Health and Bio 2016 test data. 

The en-de general domain is the WMT18 News task \cite{bojar2017findings}, with all data except ParaCrawl oversampled by 2 \cite{sennrich2017university}. We validate on newstest17 and evaluate on newstest18. We adapt first to the IWSLT 2016 TED task \cite{cettolo2016iwslt}, and then sequentially to the APE 2017 IT task \cite{apeWMT17}. 

We filter training sentences for minimum three tokens and maximum 120 tokens, and remove sentence pairs with length ratios higher than 4.5:1 or lower than 1:4.5. Table \ref{tab:data} shows filtered training sentence counts. Each language pair uses a 32K-merge source-target BPE vocabulary trained on the general domain \cite{sennrich2016subword}.  

We implement in Tensor2Tensor \cite{tensor2tensor} and use its base Transformer model \cite{vaswani2017attention} for all NMT models. At inference time we decode with beam size 4 in SGNMT \cite{stahlberg2017sgnmt} and evaluate with case-sensitive detokenized BLEU using SacreBLEU \cite{post2018call}. For BI, we use 4-gram KENLM models \cite{heafield2011kenlm}.

\begin{table}[!ht]
\centering
\small 
\begin{tabular}{ |l| l |l |}
\hline 
\textbf{Language pair} & \textbf{Domain} & \textbf{Training sentences}\\
\hline
\multirow{2}{*}{es-en}  &  Health & 586K \\
& Bio & 125K \\ 
      \hline
\multirow{3}{*}{en-de}  &  News & 22.1M \\
& TED & 146K\\ 
& IT & 11K \\
\hline
    \end{tabular}
    \caption{Corpora training sentence counts}
        \label{tab:data}
\end{table}

\subsection{Adaptive training results}
\label{sec:adaptive-training}


\begin{table}[!ht]
\centering
\small
\begin{tabular}{r|l |cc | }
    \cline{2-4}
& \textbf{Training scheme}     &  \textbf{Health} & \textbf{Bio}   \\
    \cline{2-4}
\footnotesize{1} &Health &   \textbf{35.9}      &   33.1  \\
\footnotesize{2} &Bio    &  29.6 & 36.1  \\
\footnotesize{3} &Health and Bio &  35.8 &   37.2\\
    \cline{2-4}
\footnotesize{4} &1 then Bio, No-reg  &  30.3  & 36.6 \\
\footnotesize{5} &1 then Bio, L2  & 35.1 & 37.3   \\
\footnotesize{6} &1 then Bio, EWC & 35.2 & \textbf{37.8}  \\  
    \cline{2-4}
    \end{tabular}
\caption{ Test BLEU for es-en adaptive training. EWC reduces forgetting compared to other fine-tuning methods, while offering the greatest improvement on the new domain.}

\label{tab:ewcesen}
\end{table}
\begin{table}[!ht]
\centering
\small
\begin{tabular}{  r|l| c c c    | }
    \cline{2-5}
& \textbf{Training scheme}  & \textbf{News} & \textbf{TED} & \textbf{IT} \\
    \cline{2-5}
\footnotesize{1} &News & 37.8 &  25.3  & 35.3  \\
\footnotesize{2} &TED    & 23.7 & 24.1   & 14.4\\
\footnotesize{3} &IT    & 1.6 & 1.8  & 39.6 \\
\footnotesize{4} &News and TED &38.2  & 25.5 & 35.4 \\
    \cline{2-5}
\footnotesize{5} &1 then TED, No-reg &  30.6  &  \textbf{27.0}  &  22.1   \\
\footnotesize{6} &1 then TED, L2  & 37.9 &  26.7 & 31.8  \\
\footnotesize{7} &1 then TED, EWC    & \textbf{38.3} &  \textbf{27.0} &  33.1 \\
    \cline{2-5}
\footnotesize{8} &5 then IT, No-reg&  8.0  &  6.9  &  56.3   \\
\footnotesize{9} &6 then IT, L2 & 32.3 &  22.6 & 56.9  \\
\footnotesize{10} &7 then IT, EWC   & 35.8 &  24.6 &  \textbf{57.0} \\
    \cline{2-5}
    \end{tabular}
        
\caption{Test BLEU for en-de adaptive training, with sequential adaptation to a third task. EWC-tuned models give the best performance on each domain.}
\label{tab:ewcende}
\end{table}

\begin{table*}[!ht]
\centering
\small
\begin{tabular}{|l |cc| ccc|}
\hline
\textbf{Decoder configuration}  &
\multicolumn{2}{c|}{\textbf{es-en}}  & \multicolumn{3}{c|}{\textbf{en-de}} \\

 & \textbf{Health} & \textbf{Bio} & \textbf{News} & \textbf{TED} & \textbf{IT}  \\
 \hline
 Oracle model & 35.9 & 36.1 & 37.8 & 24.1 & 39.6 \\
 \hline
Uniform & 33.1& 36.4  & 21.9 & 18.4 & 38.9\\
Identity-BI & 35.0 & 36.6 & 32.7& 25.3 & 42.6 \\
BI & 35.9& 36.5 & 38.0 & 26.1 & \textbf{44.7}\\
IS  & \textbf{36.0} & 36.8  &37.5 & 25.6 & 43.3\\
BI + IS &   \textbf{36.0} & \textbf{36.9} & \textbf{38.4} &\textbf{26.4} &\textbf{44.7} \\
\hline
    \end{tabular}
    \caption{Test BLEU for 2-model es-en and 3-model en-de unadapted model ensembling, compared to oracle unadapted model chosen if test domain is known. Uniform ensembling generally underperforms the oracle, while BI+IS outperforms the oracle.}
    \label{biesenende}
    \end{table*}
    
\begin{table*}[!ht]
\centering
\small
\begin{tabular}{|l |cc| ccc|}
\hline
\textbf{Decoder configuration} &
\multicolumn{2}{c|}{\textbf{es-en}}  & \multicolumn{3}{c|}{\textbf{en-de}} \\

 & \textbf{Health} & \textbf{Bio} & \textbf{News} & \textbf{TED} & \textbf{IT}  \\
 \hline
 Oracle model  & 35.9 & 37.8  & 37.8 & 27.0 & 57.0  \\
 \hline
Uniform & 36.0 & 36.4 &  \textbf{38.9} & 26.0 & 43.5  \\
BI + IS & \textbf{36.2} & \textbf{38.0}  & 38.7 & \textbf{26.1} & \textbf{56.4}\\
\hline
    \end{tabular}
    \caption{Test BLEU for 2-model es-en and 3-model en-de model ensembling for models adapted with EWC, compared to oracle model last trained on each domain, chosen if test domain is known. BI+IS outperforms uniform ensembling and in some cases outperforms the oracle.}
    \label{bi-individual-domains-esenende}
    \end{table*}    
    
\begin{table*}[!ht]
\centering
\small
\begin{tabular}{|l l| c |cc|}
\hline
\multicolumn{3}{|c|}{} & \multicolumn{2}{c|}{\textbf{Decoder configuration}} \\
\textbf{Language pair}&  \textbf{Model type} & \textbf{Oracle model} & \textbf{Uniform} & \textbf{BI + IS} \\
 \hline
 \multirow{3}{*}{es-en} & Unadapted & 36.4 & 34.7 & 36.6 \\
  & No-reg & 36.6 & 34.8  & -\\ 
 & EWC & 37.0 & 36.3 & \textbf{37.2}\\
 \hline
\multirow{3}{*}{en-de} & Unadapted & 36.4 & 26.8 & 38.8 \\
& No-reg & 41.7 & 31.8 & -\\
& EWC &42.1 &38.6 & \textbf{42.0} \\
\hline
    \end{tabular}
    \caption{Total BLEU for test data concatenated across domains. Results from 2-model es-en and 3-model en-de ensembles, compared to oracle model chosen if test domain is known. No-reg uniform corresponds to the approach of \citet{freitag2016fast}. BI+IS performs similarly to strong oracles with no test domain labeling.}
    \label{biesenende-totals}
    \end{table*}

We wish to improve performance on new domains without reduced performance on the general domain, to give strong models for adaptive decoding. For es-en, the Health and Bio tasks overlap, but catastrophic forgetting still occurs under no-reg (Table \ref{tab:ewcesen}). Regularization reduces forgetting and allows further improvements on Bio over no-reg fine-tuning.  We find EWC outperforms the L2 approach of \citet{barone2017regularization} in learning the new task and in reduced forgetting.

In the en-de News/TED task (Table \ref{tab:ewcende}), all fine-tuning schemes give similar improvements on TED. However, EWC outperforms no-reg and L2 on News, not only reducing forgetting but giving 0.5 BLEU improvement over the baseline News model. 

The IT task is very small: training on IT data alone results in over-fitting, with a 17 BLEU improvement under fine-tuning. However, no-reg fine-tuning rapidly forgets previous tasks. EWC reduces forgetting on two previous tasks while further improving on the target domain.

\subsection{Adaptive decoding results}

At inference time we may not know the test data domain to match with the best adapted model, let alone optimal weights for an ensemble on that domain. Table \ref{biesenende} shows improvements on data without domain labelling using our adaptive decoding schemes with unadapted models trained only on one domain (models 1+2 from Table  \ref{tab:ewcesen} and 1+2+3 from Table \ref{tab:ewcende}). We compare with the `oracle' model trained on each domain, which we can only use if we know the test domain.

Uniform ensembling under-performs all oracle models except es-en Bio, especially on general domains. Identity-BI strongly improves over uniform ensembling, and BI with $\lambda$ as in Eq. \ref{eq:lambda-bisd}  improves further for all but es-en Bio. BI and IS both individually outperform the oracle for all but IS-News, indicating these schemes do not simply learn to select a single model. 

The combined scheme of BI+IS outperforms either BI or IS individually, except in en-de IT. We speculate IT is a distinct enough domain that $p(t|x)$ has little effect on adapted BI weights. 
    
 In Table \ref{bi-individual-domains-esenende}  we apply the best adaptive decoding scheme, BI+IS, to models fine-tuned with EWC. The es-en ensemble consists of models 1+6 from Table  \ref{tab:ewcesen} and the en-de ensemble models 1+7+10 from Table \ref{tab:ewcende}. As described in Section \ref{sec:adaptive-training} EWC models perform well over multiple domains, so the improvement over uniform ensembling is less striking than for unadapted models. Nevertheless adaptive decoding improves over both uniform ensembling and the oracle model in most cases.
 
With adaptive decoding, we do not need to assume whether a uniform ensemble or a single model might perform better for some potentially unknown domain. We highlight this in Table \ref{biesenende-totals} by reporting results with the ensembles of Tables \ref{biesenende} and \ref{bi-individual-domains-esenende} over concatenated test sets, to mimic the realistic scenario of unlabelled test data. We additionally include the uniform no-reg ensembling approach given in \citet{freitag2016fast} using models 1+4 from Table  \ref{tab:ewcesen} and 1+5+8 from Table \ref{tab:ewcende}.

 Uniform no-reg ensembling outperforms unadapted uniform ensembling, since fine-tuning gives better in-domain performance. EWC achieves similar or better in-domain results to no-reg while reducing forgetting, resulting in better uniform ensemble performance than no-reg. 
 
 BI+IS decoding with single-domain trained models achieves gains over both the naive uniform approach and over oracle single-domain models. BI+IS with EWC-adapted models gives a 0.9 / 3.4 BLEU gain over the strong uniform EWC ensemble, and a 2.4 / 10.2 overall BLEU gain over the approach described in \citet{freitag2016fast}.

\section{Conclusions}
We report on training and decoding techniques that adapt NMT to new domains while preserving performance on the original domain. We demonstrate that EWC effectively regularizes NMT fine-tuning, outperforming other schemes reported for NMT. We extend Bayesian Interpolation with source information and apply it to NMT decoding with unadapted and fine-tuned models, adaptively weighting ensembles to out-perform the oracle case, without relying on test domain labels. We suggest our approach, reported for domain adaptation, is broadly useful for NMT ensembling.

\section*{Acknowledgments}
This work was supported by EPSRC grant EP/L027623/1 and has been performed using resources provided by the Cambridge Tier-2 system operated by the University of Cambridge Research Computing Service\footnote{\url{http://www.hpc.cam.ac.uk}} funded by EPSRC Tier-2 capital grant EP/P020259/1. Initial work by Danielle Saunders took place during an internship at SDL Research.

\bibliographystyle{acl_natbib}
\bibliography{refs}

\end{document}